# A Distributed Optimized Patient Scheduling using Partial Information

G. Mageshwari and E. Grace Mary Kanaga

Department of Computer Science and Engineering, Karunya University, Coimbatore, India.
magesh.5.spgm@gmail.com
grace@karunya.edu

## ABSTRACT

*A software agent may be a member of a Multi-Agent System (MAS) which is collectively performing a range of complex and intelligent tasks. In the hospital, scheduling decisions are finding difficult to schedule because of the dynamic changes and distribution. In order to face this problem with dynamic changes in the hospital, a new method, Distributed Optimized Patient Scheduling with Grouping (DOPSG) has been proposed. The goal of this method is that there is no necessity for knowing patient agents information globally. With minimal information this method works effectively. Scheduling problem can be solved for multiple departments in the hospital. Patient agents have been scheduled to the resource agent based on the patient priority to reduce the waiting time of patient agent and to reduce idle time of resources.*

## 1. INTRODUCTION

There are many kinds of software agents, with different characteristics such as mobility, autonomy, collaboration, persistence and intelligence. Software agents are used to solve distributed problem. An agent is a software entity or computer system that is capable of autonomous action in order to meet its goal [1]. Some of the characteristics of agent are reactivity, proactiveness and social ability. Reactivity means agents are able to respond to the timely changes that occur in order to satisfy its design goal. Proactiveness means agents are able to exhibit the behavior by taking initiative in order to satisfy its design goal. Social ability, this characteristic tells that agents are capable of interacting with each other agent in order to satisfy its design goal. Some additional characteristics of agents are adaptability, autonomy, collaboration, knowledgeable, mobility and persistence [2].

There are different types of agents among which personal agents, mobile agents, collaborative agents are provided the characteristics such as mobility, collaboration, intelligence or flexible user interaction. Personal agents interact directly with a user, presenting some personality or character, monitoring and adapting to the user's activities, learning the user's style and preferences, and automating or simplifying certain rote tasks Multiple software agents are suitable for use in a wide variety of application that involve in distributed computation to solve complex problems or communication between the components. Mobile agents are collecting information or perform actions when it has been sent to remote sites. Collaborative agents are communicating with groups, representing users, organizations and services. Multiple agents share the messages to interact with each other. MAS change the complex problem into simple which can be solved easily. MAS is a system composed of multiple interacting intelligent agents. It can be used to solve problems that are difficult or impossible for an individual agent.

Agent abilities are collecting data from numerous places, searching and filtering, monitoring, agent to agent negotiation and parallel computations. The agent negotiates with resource usage. Each agent owns set of tasks, requiring certain resources for a given processing time. Software agents are in the following applications such as Personal Information Management (PIM), Electronic Commerce (EC) and Business Process Management (BPM) [3]. Multi-agent systems cooperate with other agents to solve more complex tasks. Ability of multi-agent is for increasing the quality of distributed computing. Complex software systems are still difficult to implement. Simulation techniques for MAS can be utilized to test and understand the runtime behaviour of a distributed time-based system [4]. Complex problems can be changed in to individual pieces which can be easier to implement.

## 2. STATE OF ART

Dynamic load balancing considers computational load and communicational to decrease the execution time and to increase the resource utilization. In this method, monitoring phase, redistribution phase, migration phase are considered by Robson E.De Grande et al [5]. First phase is detecting load imbalance using data collection, filtering and selection of data. Second phase determine possible migration moves. Then transfer from an overloaded resources to an under loaded resources. Computational load imbalance is avoided when the load is redistributed for communication purpose. Third phase consider two phase migration technique. It is used in redistribution scheme to minimize the migration latency and enable better balancing reactivity. According to Feng Zhang et al [6], price based multi-agent scheduling and coordination is used to determine the timeliness of scheduling in a distributed environment. This framework applied in the inventory system. It is scalable. The number of asset is varied to reflect the different problem scales. But the agents cannot retrieve the information automatically. Intra-organization as well as inter-organization coordination needs to be improved. Decision process is used for non-emergency admission planning in the hospital activities.

According to Luiz Guilherme Nadal Nunes et al [7], Markov Decision Process (MDP) is used to guide more efficient decision process to balance the non-emergency patients admission and available resource utilization in the hospital. Medical Path Coordination (MedPaCo) is a coordination mechanism which is used to minimize waiting time of the patients [8]. The idea of this coordination mechanism is that the patient agents are buying the time slots of resources for the needful treatments and examinations of the patients. Adaptive dynamic approach is used to schedule the resources automatically. It minimizes the access time of all the resources in the hospital. According to Ivan B.Vermeulen et al [9], adaptive urgent scheduling schedules the emergency patients on time. It provides flexible resource allocation for emergency patients. Health Examination Scheduling Algorithm (HESA) is used to allot the examinee to any free resources or doctor based on round robin algorithm. This method assigns the priority for the patient based on their treatment. But it does not consider, when the examinees are absent, equipment breakdown or longer than expected processing times in the resources [10]. Multi-agent Pareto Appointment Exchanging (MPAEX) system is used to exchange appointment schedules such that no patient is worse off. It uses negotiation approach to improve the condition of patient which is called Pareto improvement [11], [12]. Partial plan is developed for the patient with less number of tasks or activities. It can exchange single activity. Workload distribution has been scheduled using theil index in MPAEX approach. Generalized Partial Global Planning (GPGP) with new resource constrained mechanism is used to improve the throughput of the hospital unit and reduces the stay time of the patient [13]. Bidding is used in GPGP coordination mechanism to buy the resource timeslot for the patient agent. When two tasks need to use the same resource at a time, highest priority will be given to the task which is arrived the hospital first. New coordination mechanism is introduced with GPGP approach for handling mutually exclusive resource relationship [14].

Dynamic changes of patient scheduling problem is solved by agent based, patient-centred approach [15]. This approach is used to avoid the resource conflicts. Medical Path Agent (MedPAge) shows the roles of the agent, dynamic interaction and the coordination object identification [16]. MedPAge project handles the complicated tasks through the market mechanism which is supported by software agent in the hospital. To improve the current scheduling situation, agent interaction protocol is used. The agents interact with each other. This mechanism maximizes the patient's own utility. Distributed multi-agent based approach is implemented in the Medical Path Coordination (MedPaCo) for coordination of patient agents [17]. The aim of this coordination mechanism is to reduce the overall waiting time of the patients. This approach implements the patients and hospital resources as software agents through the concept of MedPaCo [18]. This mechanism is based on auction or bidding process which consists of certain phases. These phases make the patient to get highest utility from a timeslot by paying highest amount for that timeslot. This coordination mechanism cannot handle multiple preferences of the patient agents.

## 3. PROBLEM DESCRIPTION

Scheduling is difficult to implement in the hospital because of the dynamic changes. Medical data management systems and Decision support systems are analyzed by agent based approach. Medical data management systems are focused on the processing of medical data and retrieval of data such as electronic health record. Decision support systems approaches are aimed to help the execution of healthcare processes such as treatments [4]. According to ReddyMC et al. [19], [20] challenges are found in the hospital from various analysis. First challenge is ineffectiveness of current information and communication technologies. Cell phones, 2-way radios are the communication technologies which is failed to work effectively because of mismatch frequencies. Clinical/nonclinical department interaction makes inappropriate patient transfer. Clinical/clinical department interaction increases the waiting time of the patient in a department. Lack of complete and accurate information shared between the departments because of dynamic changes in the hospital. Without appropriate information, inpatient access department may make inappropriate assignments as well as deal with temporary bottleneck problem in the patient flow by holding up with resources. Breakdowns in information flow include the loss of patient information transferring the patients to wrong locations due to misrepresentation of patient's information. Master health checkup includes blood test, urine test along with Electro cardio gram (E.C.G), ultra scan, X-ray, thyroid function test, dental consultation and eye consultation. In master health checkup, patients can undergo any checkup. When a particular resource gets overcrowded the patients can go any free resources because there is no constraint. New method has been proposed in this paper to transfer the group of patients from overcrowded resources to free resources.

## 4. THE PROPOSED DISTRIBUTED PATIENT SCHEDULING WITH GROUPING METHOD

This section describes the architecture of patient scheduling, flow chart and the algorithm of the proposed method.

### 4.1. Architecture Overview

Agent scheduling process is developed by Qingqi Long et al [21] for dynamic load balancing in agent based simulation. Each agent executes the simulation by message communication. Based on this agent scheduling, this paper propose a method to schedule the patients to the resources in the hospital. The architecture of patient scheduling is shown in Figure 4.1.

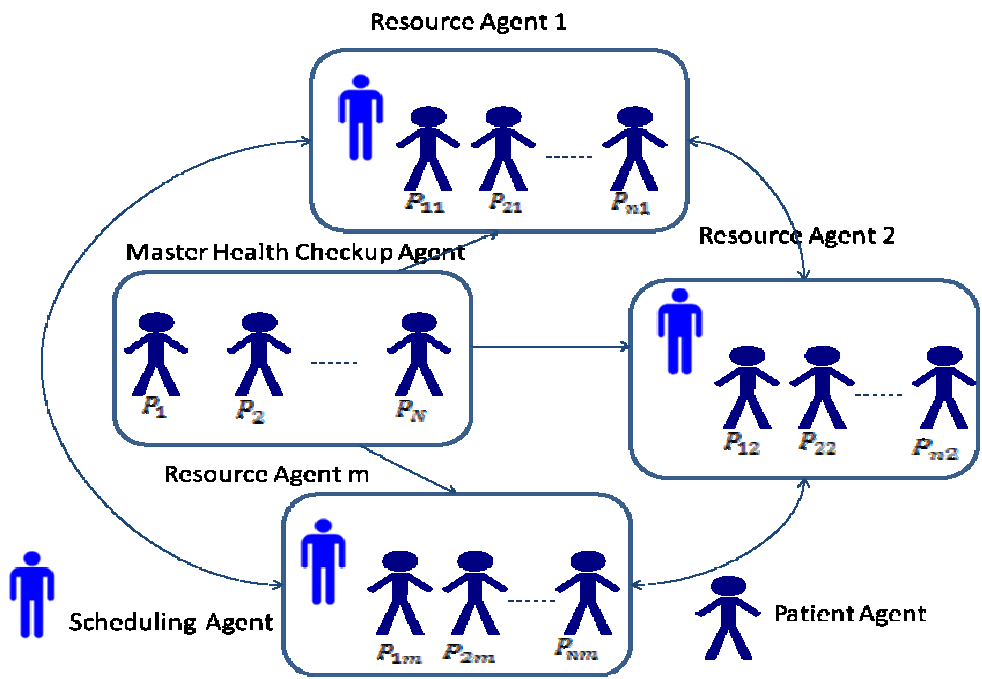

Figure 4.1 Architecture of Patient Scheduling

Master Health Checkup Agent (MHCA) is in charge of receiving, processing the Patient Agent (PA) information. It also maintains the count of patients. Resource Agents (RA) send and receive the requests from other resources. Responsibility of MHCA is to send a PA to the RA. It is also in charge of monitoring PA and RA information and agent communication. In addition, each resource is equipped with a Scheduling Agent (SA) that is in charge of triggering patient scheduling procedure, selecting the migrated agent sets and target containers, and moving the agents. The SA is the centre of patient scheduling and migration in this simulation platform. As each resource has a SA, the scheduling is done in a distributed way.

When a paricular resource gets overcowded it needs to migrate the exceeding number of patients to another resource in order to reduce the waiting time of the patient and to improve the resource utilization. First the overcrowded resource will send request to neighborhood resource which response if it has enough space to treat that patient. Otherwise it will ignore the request. After that overcrowded resource will send request to another resource until overcrowded will be accepted by any neighborhood resource. Using migration set approach, minimizes the number of communication and grouping the patient agent based on their treatment. Resource agents will interact with each other. Each resource agent does not have any information about the patient agent in another resource agent. Figure 4.2 show that how a set of patients migrated from overcrowded resource agent to neighbourhood resource agent.

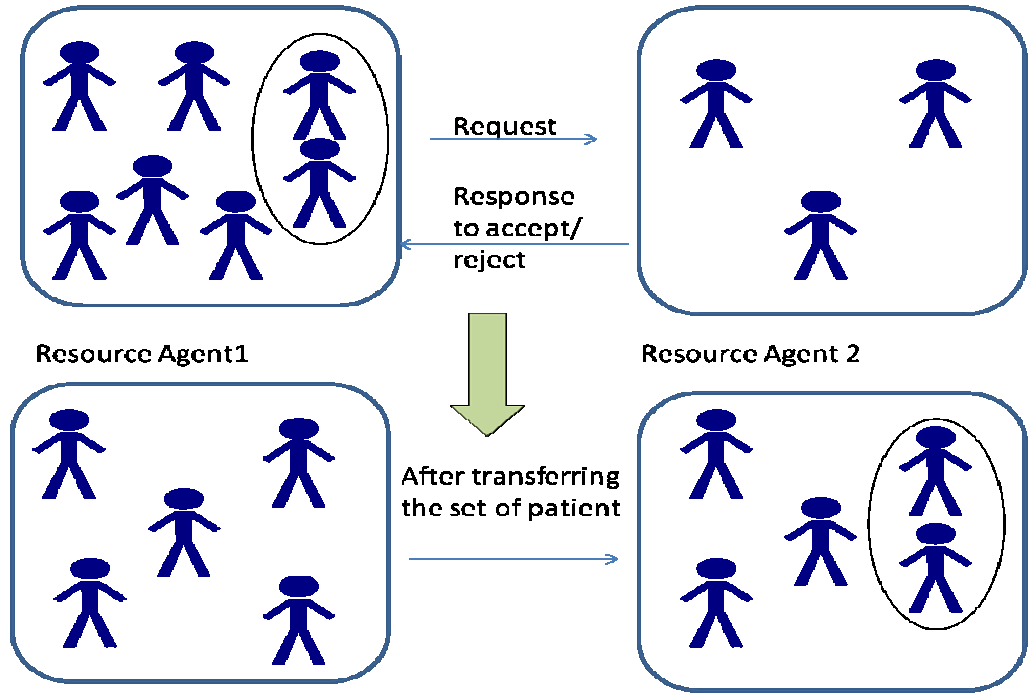

Figure 4.2 Set of Patient Transferred using Migration Set Approach

## 4.2. Algorithm and Flow Diagram of Proposed Method

Table 4.1 shows the notations which are used in this method. The proposed method reduces the waiting time of the patient and improves the patient satisfaction.

Table 4.1 Notations

| Parameter | Description |
|---|---|
| $RA_i$ | Overcrowded resource agent where i varies from 1 to n |
| $RA_{i+1}$ | Neighbourhood resource agent where i varies from 1 to m |
| $CRA_{i+1}$ | Current Capacity of i+1th resource agent |
| $ThRA_{i+1}$ | Fixed Capacity of i+1th resource agent |
| $G_i$ | Group of patients in ith resource agent |
| $E_i$ | Exceeded number of patients from ith resource agent |
| $P_{ij}$ | Patient agent j in resource agent i where i varies from 1to m where j varies from 1 to ni |

Figure 4.3 shows the flow diagram of transferring set of patients. $RA_i$ represents the overcrowded resource agent. $RA_{i+1}$ represents the neighborhood resource agent without crowd. When a particular resource agent gets overcrowded then that will send request to another neighborhood resource agent. It will check whether it has enough capacity to do treatment for another set of patient. If it has space it will accept. Otherwise it will reject the request from overcrowded resource agent. Group the patients from exceeded number of patient based on their priority, new arrival time and next task. That group of patients will be considered as a set $G_i$.

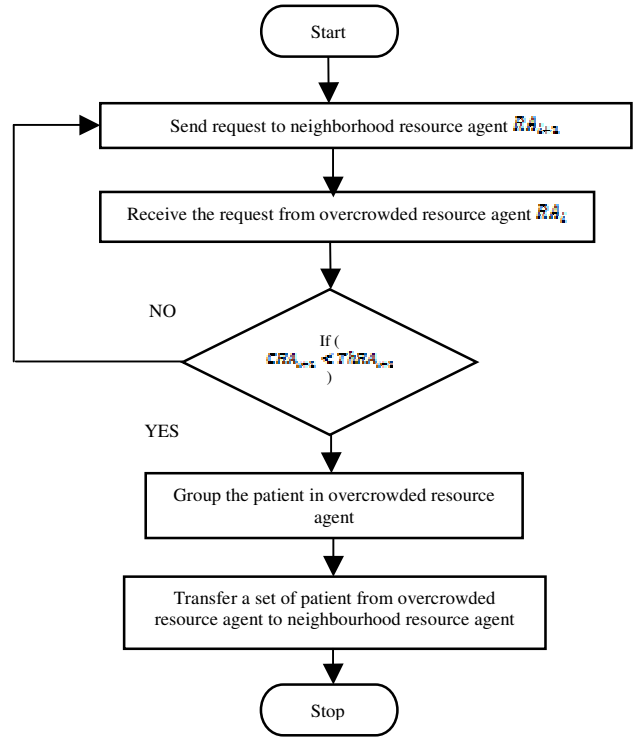

Figure 4.3 Flow Diagram for Transferring Group of Patients

---

Algorithm1: Distributed Optimized Patient Scheduling using Partial Information

---

$RA_i$ sends request to $RA_{i+1}$
**If** ($CRA_{i+1} < ThRA_{i+1}$) **then**
    Accept request
**Else**
    Reject request
**End if**
**For** ($P_{ij} \in E_i$) **do**
    Group the set $G_i$ (i.e) set of patient in exceeded number of patient
    Move $G_i$ to $RA_{i+1}$
    Remove $G_i$ from $RA_i$
**End for**

---

Figure 4.4 Distributed Optimized Patient Scheduling using Partial Information

The pseudo-code of the proposed algorithm is given in Figure 4.4. When a particular resource agent gets overcrowded then that will send request to another neighborhood resource agent. It will check whether it has enough capacity to do treatment for another set of patient. If it has space it will accept. Otherwise it will reject the request from overcrowded resource agent.

Group the patients from exceeded number of patient in the overcrowded resource agent based on their priority, new arrival time and next task. That group of patients will be considered as a set $G_i$. Finally, patients are transferring from overcrowded resource agent to neighborhood resource agent. Set of patients will be moved to another resource agent. Then migrated agent will be removed from overcrowded resource agent. It will be repeated until the entire overcrowded patient is scheduled. This method DOPSG has been used to search for optimized scheduling scheme, including migrated agent sets.
The steps of the proposed algorithm are as follows,

*Step1:* $RA_i$ send request to $RA_{i+1}$ with id of $RA_i$ and exceeded number of patients ($E_i$).
*Step2:* If ($CRA_{i+1} < ThRA_{i+1}$) then accept the request. Otherwise go to step1.
*Step 3:* When a patient belongs to exceeded number of patient then that patient has to be grouped based on priority, new arrival time, and next task. Group of patients considered as a set $G_i$.
*Step4:* Move $G_i$ to $RA_{i+1}$ and remove $G_i$ from $RA_i$ .
*Step5:* Repeat until all exceeded patients transfer from overcrowded resource agent.
*Step 6:* Stop.

## 5. EXPERIMENTAL RESULTS

To implement this scheduling method for distributed patients with grouping, hardware requirements and software requirements are used. Hardware Requirements are Intel® Core (TM) 2 Duo CPU T8100 @ 2.10 GHz processor and 32- bit operating system are used. Software Requirements are JDK 1.6 java software used with Java Agent Development Framework (JADE) package. Three resources and fifty patients are created to do the scheduling. Initially create the patient agent in MHCAgent by retrieving from database. Each patient agent have unique agent id which will be generated while creating agent. Created patient agent should send to resource agent to do their treatment. Through agent communication one resource agent sends the message to another resource agent. After receiving the responses, group of patient agents are moved from one resource agent to other resource agent. It minimizes the number of communication. Then scheduling has been done with grouping.

### 5.1. Performance Metrics

The performance of the multi-agent system can be evaluated by following metrics. The aim of patient scheduling problem is to minimize the patient waiting time and idle time of resources in the hospitals. Performance metrics are useful in evaluating the objectives of patient scheduling is given below:

***Maximum completion time (Cmax)***
This refers to difference between the starting time and finishing time of a sequence of jobs or tasks. In terms of patient scheduling, it refers to the difference between when the patient checks in and the time taken to complete all the tasks in their diagnosis.
$$Cmax = \text{Maximum of end time}$$

***Maximum tardiness (Tmax)***
It refers to the lateness of a job which is a value equal to the completion time minus the due time. This metric gives the worst case scenario of how long a patient might be delayed in the worst case.
$$Tmax = \text{Maximum of Tardiness}$$
$$Tardiness = Cmax - Duetime$$
$$Duetime = Arrivaltime + Totalprocessing\ time$$

*Total completion time (∑Cj)*
It represents the total amount of time utilized in the schedule. For patient scheduling we want to minimize this metric to minimize the patient stay in the hospital.
∑*Cj*= sum of completion time

*Total tardiness (∑Tj)*
It measures the overall total tardiness of all the patients.
∑Tj= sum of tardiness

*Total weighted completion time (∑WjCj)*
This metric includes the priority of the patient when calculating the total flow time.
∑WjCj= sum (weight*completiontime)

*Total weighted tardiness (∑WjTj)*
This is the main metric is used to evaluate the patient scheduling algorithm to indicate whether the patient waiting time is minimized.
∑WjTj= sum (weight*tardiness)

A good scheduling technique will effectively minimize the values of all these metrics. This proposed approach also aims to minimize the weighted tardiness and the weighted completion time.

## 5.2. Performance Analysis

Performance analysis has been shown in the graph. Figure 5.1 shows the comparison of total completion time of the patients of proposed method with traditional scheduling techniques like First Come First Serve (FCFS), Weighted Shortest Processing Time (WSPT), Distributed Optimized Patient Scheduling (DOPS) and Distributed Optimized Patient Scheduling with Grouping (DOPSG).

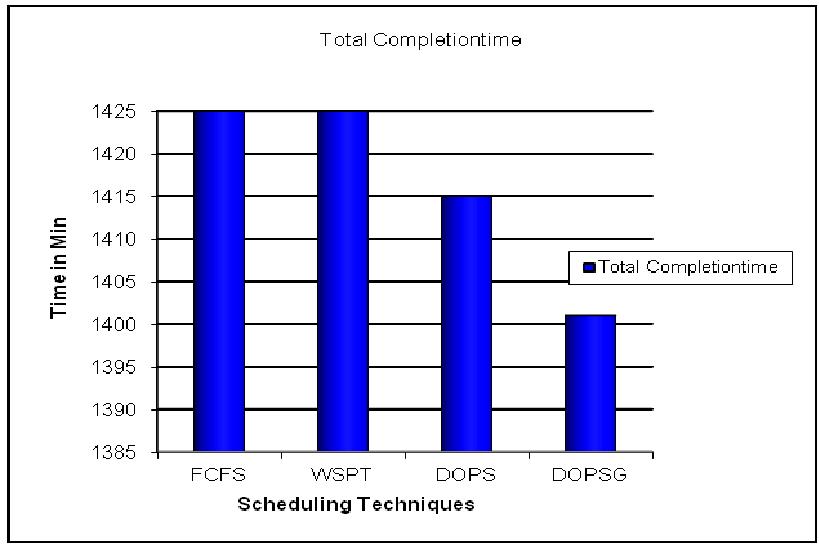

Figure 5.1 Comparison of Total Completion Time

DOPSG minimizes number of communication than the other scheduling. The total completion time of DOPSG is reduced by 1% when compared to DOPS, 1.7% when compared to WSPT and 1.7% with respect to FCFS. Figure 5.2 shows total tardiness of the patient in various scheduling techniques. The total tardiness of DOPSG is reduced by 9% when compared to DOPS, 12% when compared to WSPT and 12% with respect to FCFS.

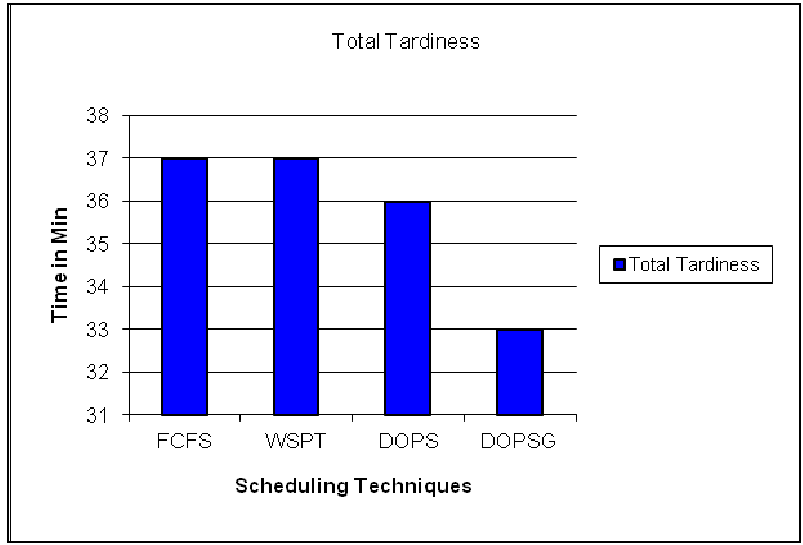

Figure 5.2 Comparison of Total Tardiness

Figure 5.3 shows the total weighted completion time of the patients in various scheduling approaches. The total weighted completion time of DOPSG is reduced by 0.8% when compared to DOPS, 1.1% when compared to WSPT and 1.2% with respect to FCFS.

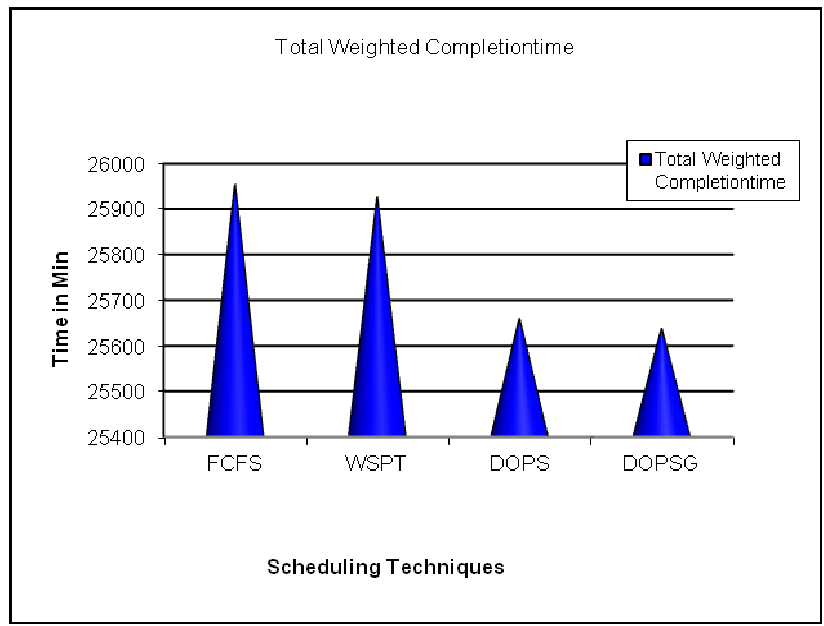

Figure 5.3 Comparison of Total Weighted Completion Time

Figure 5.4 shows total weighted tardiness of the patient in various scheduling methods. The total weighted tardiness of DOPSG is reduced by 5% when compared to DOPS, 37% when compared to WSPT and 41% with respect to FCFS.

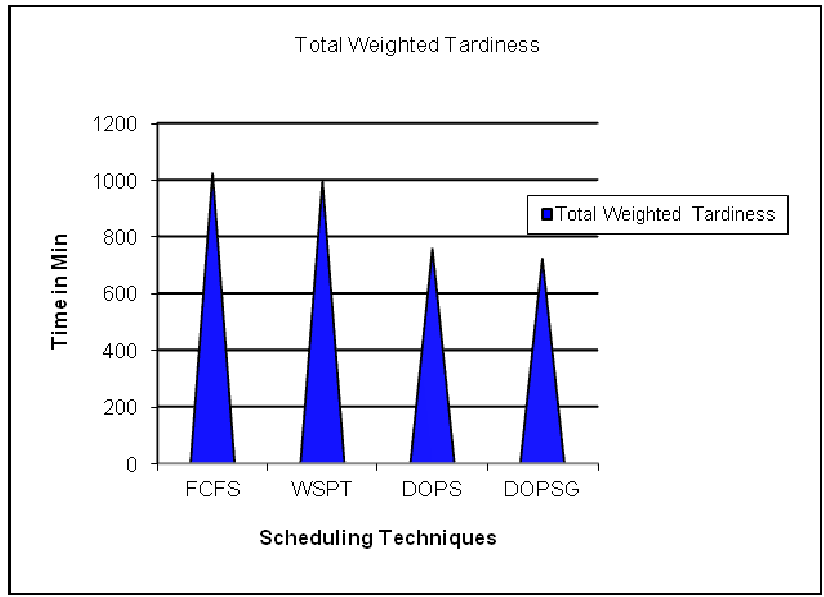

Figure 5.4 Comparison of Total Weighted Tardiness

## 6. CONCLUSIONS

A distributed approximate optimized patient scheduling algorithm with partial information has been proposed and implemented in this paper. First, this algorithm does not need to know all the information of departments where agents are, so scheduling can be carried out with partial information. Secondly, since this algorithm is distributed, it can not only avoid bottleneck problem and potential failure problem caused by the centralized mode, but also improve resource utilization. Thirdly, this algorithm is approximately optimized. This proposed algorithm is used to reduce the waiting time of patients and thus improving the health state of patient and to improve resource utilization.